\documentclass[pmlr]{jmlr} 

\usepackage{booktabs} 
\usepackage{graphicx} 
\usepackage{url}      
\usepackage{amsmath}  
\usepackage{algorithm2e} 
\usepackage{caption}
\usepackage{subcaption}
\selectfont
\usepackage{parskip}

\usepackage{fancyhdr}

\makeatletter
\def\@jmlrproceedings{}
\def\@jmlrpages{} 
\makeatother

\title{Physics-based phenomenological characterization of cross-modal bias in multimodal models\\[1.5em]}

\makeatletter

\providecommand{\Name}[1]{\textbf{#1}}
\providecommand{\Email}[1]{\hfill\texttt{#1}}
\providecommand{\addr}[1]{\\\textit{#1}}

\providecommand{\AND}{\par\vspace{0.8em}}

\makeatother

\makeatletter
\newcommand{\AuthorNamesOnly}{%
  Hyeongmo Kim, Sohyun Kang, Yerin Choi, Seungyeon Ji, Junhyuk Woo, Hyunsuk Chung, Soyeon Caren Han, Kyungreem Han%
}
\ifdefined\HCode
  \renewcommand{\author}[1]{\gdef\@author{\AuthorNamesOnly}}
\fi
\makeatother

\author{\Name{Hyeongmo Kim}\thanks{These authors contributed equally.} \Email{momo1180@kist.re.kr}\\
    \addr Brain Science Institute, Korea Institute of Science and Technology, Seoul, Republic of Korea \\
    Department of Physics and Astronomy, Seoul National University, Seoul, Republic of Korea
  \AND
  \Name{Sohyun Kang}$^*$ \Email{sohyunkang@kist.re.kr}\\
  \addr Brain Science Institute, Korea Institute of Science and Technology, Seoul, Republic of Korea
  \AND
  \Name{Yerin Choi}$^*$ \Email{yerin.choi7@kist.re.kr }\\
  \addr Brain Science Institute, Korea Institute of Science and Technology, Seoul, Republic of Korea
  \AND
  \Name{Seungyeon Ji}$^*$ \Email{syji@kist.re.kr}\\
  \addr Brain Science Institute, Korea Institute of Science and Technology, Seoul, Republic of Korea \\ Department of Computer Science and Engineering, Korea University, Seoul, Republic of Korea
  \AND
  \Name{Junhyuk Woo}$^*$ \Email{wjh601@kist.re.kr }\\
  \addr Brain Science Institute, Korea Institute of Science and Technology, Seoul, Republic of Korea
  \AND
  \Name{Hyunsuk Chung}$^*$ \Email{hyunsuk.chung.1@unimelb.edu.au }\\
  \addr The University of Melbourne, Melbourne, Australia
  \AND
  \Name{Soyeon Caren Han}$^{*,}$\thanks{Corresponding authors.} \Email{caren.han@unimelb.edu.au}\\
  \addr The University of Melbourne, Melbourne, Australia
  \AND
  \Name{Kyungreem Han}$^{*,\dagger}$\Email{khan@kist.re.kr}\\
  \addr Brain Science Institute, Korea Institute of Science and Technology, Seoul, Republic of Korea \\
  University of Science and Technology KIST School, Seoul, Republic of Korea
}

\begin{document}

\maketitle

\vspace{-1em}
\begin{abstract}
The term `algorithmic fairness' is used to evaluate whether AI models operate fairly in both comparative (where fairness is understood as formal equality, such as “treat like cases as like”) and non-comparative (where unfairness arises from the model’s inaccuracy, arbitrariness, or inscrutability) contexts. Recent advances in multimodal large language models (MLLMs) are breaking new ground in multimodal understanding, reasoning, and generation; however, we argue that inconspicuous distortions arising from complex multimodal interaction dynamics can lead to systematic bias. The purpose of this position paper is twofold: first, it is intended to acquaint AI researchers with phenomenological explainable approaches that rely on the physical entities that the machine experiences during training/inference, as opposed to the traditional cognitivist symbolic account or metaphysical approaches; second, it is to state that this phenomenological doctrine will be practically useful for tackling algorithmic fairness issues in MLLMs. We develop a surrogate physics-based model that describes transformer dynamics (i.e., semantic network structure and self-/cross-attention) to analyze the dynamics of cross-modal bias in MLLM, which are not fully captured by conventional embedding- or representation-level analyses. We support this position through multi-input diagnostic experiments: 1) perturbation-based analyses of emotion classification using Qwen2.5-Omni and Gemma 3n, and 2) dynamical analysis of Lorenz chaotic time-series prediction through the physical surrogate. Across two architecturally distinct MLLMs, we show that multimodal inputs can reinforce modality dominance rather than mitigate it, as revealed by structured error-attractor patterns under systematic label perturbation, complemented by dynamical analysis. 
\end{abstract}

\begin{keywords}
MLLM, cross-modal bias, multi-oscillator model, physics-based explanation
\end{keywords}

\section*{Background and Introduction}

Recent advances in large language models that enable multimodal understanding, reasoning, and generation will open new possibilities for versatile human-machine collaboration, provided that safety and ethical concerns are properly addressed. Multimodal large language models (MLLMs) have transformed how we interact with machines—they utilize not just natural language (text), but also audio, images, and video, just like humans do daily \citep{mllm-survey,mm-llms}. Combining multiple modalities into a single AI model seems theoretically plausible because it could more closely mirror how humans integrate (in the multimodal association area in the cerebral cortex \citep{amaral2012_cortex}) multimodal information (e.g., vision, hearing, smell and touch) for the higher-level cognitive functions. However, recent empirical evidence suggests that such multimodal integration does not necessarily lead to fair or robust decision-making, and can instead introduce subtle but systematic bias that is not captured by aggregate performance metrics \citep{text_takes_over}.

\quad Modern MLLMs involve a series of steps to associate heterogeneous modalities with their distinct data structures properly and to generate rich outputs: data encoding, feature projection, feature fusion, cross-modal processing, multimodal output decoding. To summarize: The real-world data across different modalities are encoded into machine-understandable features (e.g., tokenization for written/spoken language to produce semantic word embeddings), and then high-level features are produced. Through a projection step, the abstract high-level features are meaningfully associated into a shared embedding space as a numerical vector representation. Next, feature fusion is performed to form a unified multimodal representation, such as via a cross-attention mechanism. The cross-modal processing step develops subtle cross-modal dependencies from the fused multimodal representation by leveraging the transformer architecture, which stacks self-attention and feedforward layers to refine context within the same modality and cross-attention to integrate distinct modalities. The decoding step produces an output for complex multistep tasks, ensuring the model’s understanding, reasoning, and generation are delivered in a usable format. While this transformer dynamics enables impressive zero-shot and few-shot performance, it can also give rise to subtle failure modes that are not reflected in aggregate accuracy.

\quad Besides the positive and promising aspects, there are also challenges related to their black-box-like nature, which can cause unexpected, seemingly unavoidable distortions, such as performance drops and bias generation. Multimodal models tend to exhibit significant modality bias, with decisions determined primarily by a single modality \citep{Gao2024ModalityBias,MISA}, while undominant modalities offer little useful information and may even introduce noise that interferes with the model's decision-making. Consistent with this observation, recent empirical results from studies on modality bias in multimodal intent detection show that, in certain task settings, unimodal text-only LLMs can outperform multimodal LLMs, indicating that modality bias can lead to measurable performance decline \citep{text_takes_over}. For the medical domain, empirical evidence indicates that medical MLLMs often perform at levels similar to, or even lower than, unimodal LLMs in both general and medical fine-tuning contexts \citep{tu2024_medical,dai2025_medical,buckley2023_medical}. For example, \cite{dai2025_medical} examine multimodal diagnosis using paired chest X-ray images and clinical text descriptions, and find that both general-purpose multimodal models (e.g., GPT-4-class and Gemini-class systems) and medically fine-tuned models (e.g., LLaVA-Med) tend to rely predominantly on textual clinical information, while visual features from X-ray images contribute little to predictions and can even degrade performance. Similarly, \cite{buckley2023_medical} study multispecialty case-based diagnosis spanning dermatologic, radiologic, ocular, and oral conditions, and demonstrate that these multimodal foundation models can often reproduce image-based diagnostic performance using text-only inputs, suggesting that predictions are driven by textual priors rather than genuine visual reasoning.

\quad We hypothesize that 1) these counterintuitive performance variations arise primarily from distortions in transformer dynamics, and 2) traditional cognitivist symbolic accounts (e.g., embedding- or representation-level analyses) or metaphysical approaches cannot properly characterize them. The undesired yet inconspicuous distortions in numerical representations may occur due to biased transformer dynamics that reflect self-attention and feedforward operations (within the same modality) and cross-attention (across multiple modalities)—eventually leading to inadequate development of cross-modal dependencies. Such failures are not merely performance issues; they directly relate to algorithmic fairness, particularly in non-comparative settings where arbitrariness and inscrutability arise even without explicit group comparisons. As such, this paper presents the development of a surrogate physics-based model that captures the core features of transformer dynamics, including its semantic network structure and self- and cross-attention mechanisms to analyze the dynamics of cross-modal bias in MLLMs. We begin with a diagnostic analysis of emotion classification on Qwen2.5-Omni \citep{qwenomni} and Gemma 3n \citep{gemma3n} models. A dynamic analysis of the multi-oscillator surrogate model for the Lorenz chaotic time-series prediction task follows this.

\section*{Diagnostic Analysis of Multimodal Large Language Models}

\subsection*{Experimental Setup}
We conduct perturbation-based analyses of emotion classification using CREMA-D (Crowd-sourced Emotional Multimodal Actors Dataset) \citep{cao2014_CREMAD} with two architecturally distinct MLLMs, Qwen2.5-Omni and Gemma 3n. We compare three input conditions: 1) combined video (face) and audio (voice) inputs, 2) video (face)-only (audio replaced with silence), and 3) audio (voice)-only (video frames replaced with blank placeholders) to identify modality-dependent error attractor patterns.
\begin{table}[!b]
\centering

\begin{tabular}{|p{0.97\linewidth}|}
\hline
\small
\texttt{Below is a list of possible emotions: \{emotion\_list\}.} \\
\texttt{Your task: Recognize dominant emotion expressed in the \{modality\_list\}.} \\
\texttt{Return the result in valid JSON array format as follows:} \\ \texttt{[\{"emotion": ""\}].} \\
\texttt{Do not output anything else (no extra commentary).} \\
\hline
\end{tabular}
\caption{The prompt for the zero-shot emotion classification task. The \texttt{\{modality\_list\}} field is set to "face and voice" for combined video–audio inputs, "face" for video-only inputs (with audio replaced by silence), and "voice" for audio-only inputs (with video frames replaced by blank placeholders).}
\label{tab:prompt}
\end{table}
The CREMA-D dataset includes videos of actors expressing specific emotions (i.e., happy, neutral, sad, angry, disgust, and fear) against a green screen with synchronized audio. It offers a multi-layered annotation structure: the intended emotion (the emotion actors were instructed to display, single-label) and perceived emotions labeled by humans across three conditions: multimodal-perceived (original video with audio), visual-perceived (video-only without audio), and audio-perceived (audio-only without video). Perceived emotion labels are multiple when multiple emotions receive equal vote counts. Each sample was rated by an average of 9.8 annotators, with over 95\% of samples receiving at least 8 independent evaluations. The dataset includes 7,442 samples frm 91 actors, all of which were used in our experiments.

\quad We conduct zero-shot classification experiments on Qwen2.5-Omni and Gemma 3n to examine inherent model bias independent of fine-tuning. The prompt used zero-shot classification for both models is shown in Table \ref{tab:prompt}.

\begin{figure}[ht]
\centering

\begin{minipage}{0.46\textwidth}
\centering
\includegraphics[width=\textwidth]{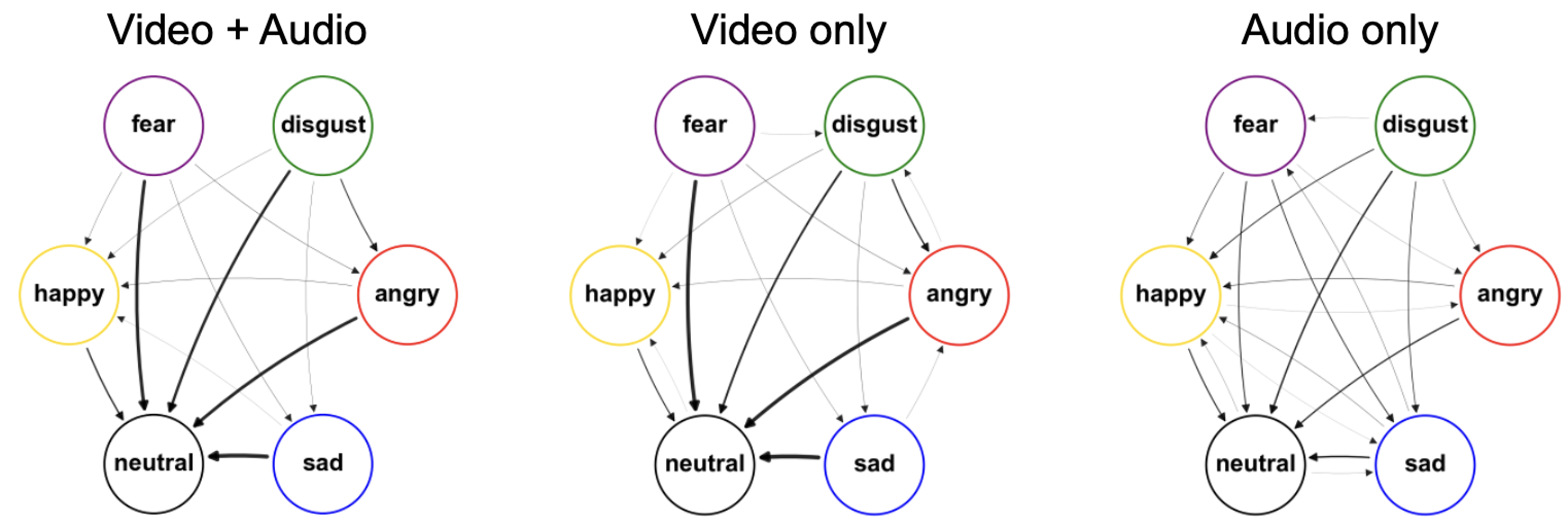}
\\
\small (a) Qwen2.5-Omni
\end{minipage}
\hfill
\begin{minipage}{0.46\textwidth}
\centering
\includegraphics[width=\textwidth]{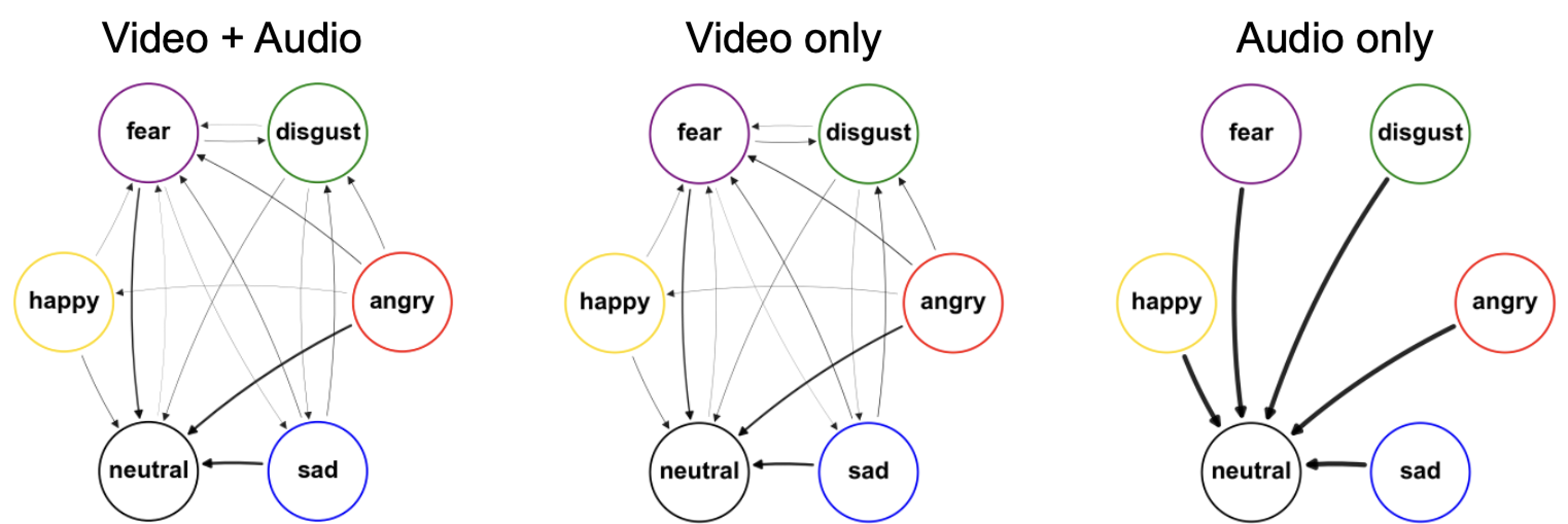}
\\
\small (b) Gemma 3n
\end{minipage}

\caption{Error-attractor structures under perturbations of emotion labels in multimodal large language models.
Directed graphs visualize incorrect emotion classifications on the CREMA-D dataset under three input conditions:
Face (Video) + Voice (Audio), Face (Video) only, and Voice (Audio) only.
Nodes represent six emotion labels (happy, neutral, sad, angry, disgust, fear), and directed edges depict erroneous mappings from intended labels to predicted labels.}
\label{fig2}
\end{figure}




\begin{figure}[t]
\centering
\includegraphics[scale=0.34,angle=0]{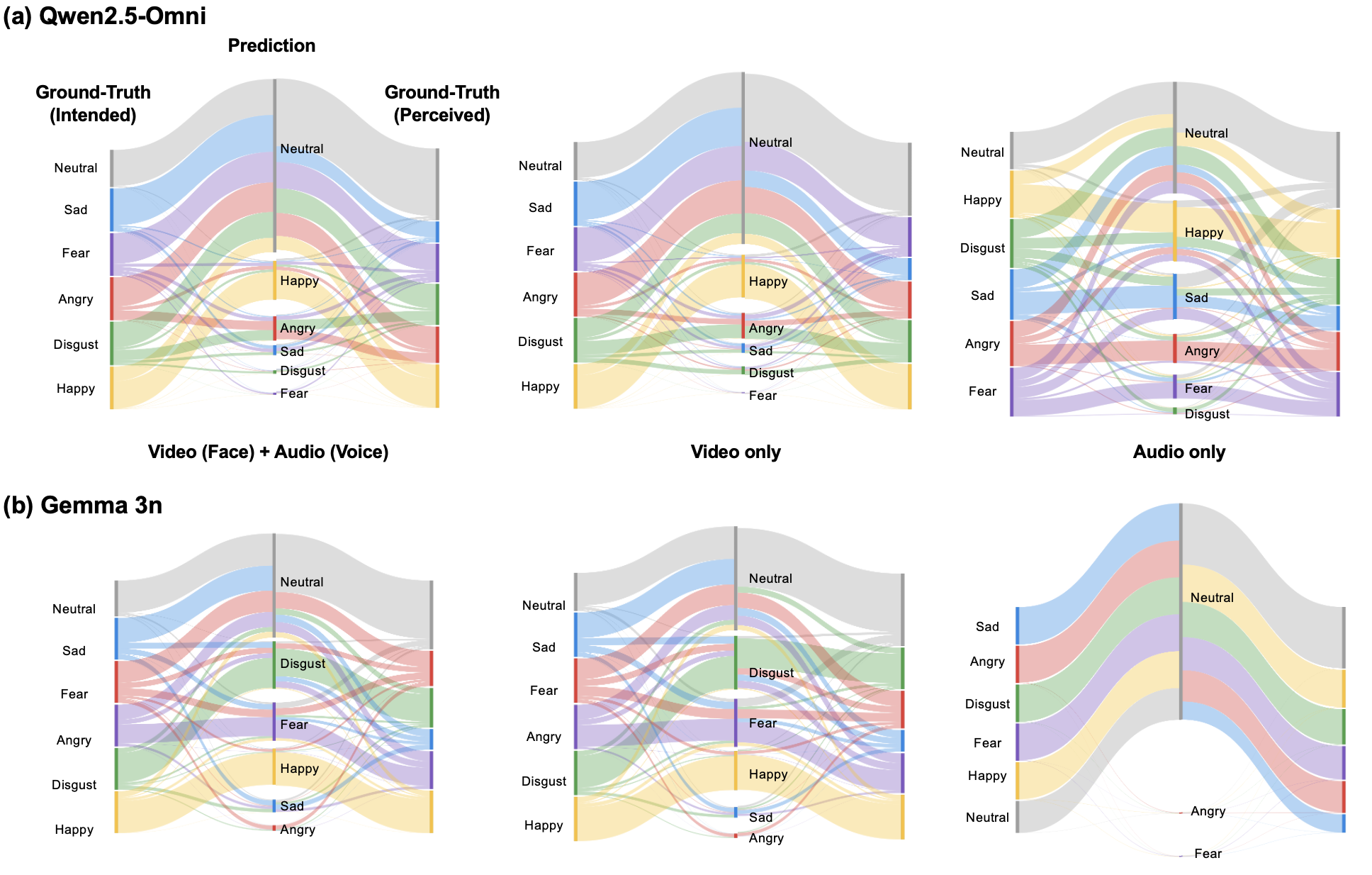}
\caption{\label{fig3}
Sankey diagram displaying emotion predictions in Qwen2.5-Omni. The width of each flow indicates the number of samples assigned to each mapping: 1) from the intended label (left) to the model prediction (center), and 2) from the perceived label (right) to the model prediction (center).
}
\end{figure}

\subsection*{Emotion Label Perturbation Analysis}
\subsubsection*{Hierarchical bias revealed through label perturbation}
Before applying label perturbations, we first analyze model behavior in a full-label setting, where all emotion categories are available for selection. Figure \ref{fig2} visualizes incorrect emotion classifications on the CREMA-D dataset using a directed graph representation for (a) Qwen2.5-Omni and (b) Gemma 3n. Nodes correspond to the six emotion labels (happy, neutral, sad, angry, disgust, fear) in CREMA-D, and directed edges represent erroneous mappings from intended labels to model predictions. Edge thickness is proportional to the number of corresponding misclassifications, with low-occurrence transitions suppressed for clarity. Self-loops (correct predictions) are omitted to focus exclusively on bias patterns under failure. 

\quad Based on the intended emotion annotation, both Qwen2.5-Omni and Gemma 3n exhibit significant biases in their model predictions. The bias patterns vary depending on the model structure and input modality, whether it involves combined video (face) and audio (voice), video-only, or audio-only.  

\quad The bias patterns are further examined using a Sankey diagram, shown in Figure \ref{fig3}. The diagram illustrates the bias hierarchy of the Qwen2.5-Omni model by mapping actors’ intended emotions (left) and third-party perceived emotions (right) to the model-predicted emotions (center). For both intended and perceived emotion annotations, predictions consistently exhibit a clear hierarchical bias. The emotion predictions are most strongly concentrated in the neutral category, yet the detailed bias dynamics differ across input modalities. Notably, the mapping pattern in the Face+Voice case closely resembles that in the Face-only condition, while the Voice-only condition displays a different pattern. 

\quad We then apply a systematic prompt-based perturbation strategy \citep{khanmohammadi2025_emotion}, in which models are instructed to perform classification while being explicitly prohibited from selecting subsets of emotion labels (removing one, two, three, or four labels at a time). This procedure reveals a hierarchical structure of error attractors: when preferred labels are removed, models consistently fall back to secondary or tertiary choices rather than redistributing errors uniformly.

\begin{figure}
\centering
\includegraphics[scale=0.31,angle=0]{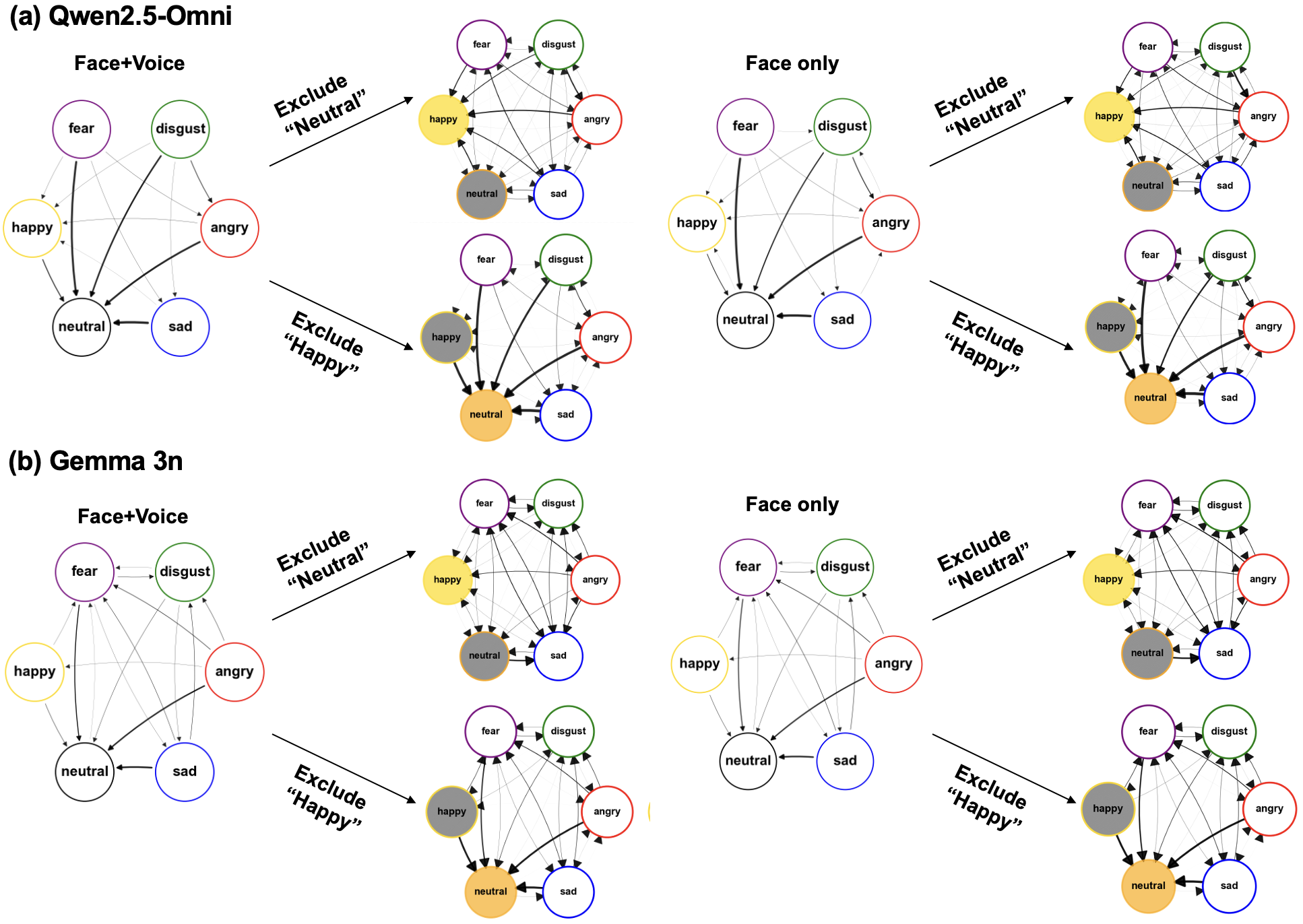}
\caption{\label{fig4}
Illustration of the prompt-based label perturbation strategy
}
\end{figure}

\quad Across both models (Fig. \ref{fig4}), erroneous predictions are highly structured rather than random. This suggests that misclassification patterns encode implicit preference hierarchies over emotion labels \citep{dong2020_emotion}, which can be interpreted as a form of reinforcement bias in the output space. Such hierarchies are informative precisely because they are exposed only when the model fails, and are therefore invisible to standard accuracy-based evaluations. 

\quad In Qwen2.5, Neutral consistently emerges as the dominant attractor across perturbation conditions. When Neutral is excluded from the prompt under Face+Voice or Face-only inputs, Happy becomes the next dominant attractor; excluding Happy, in turn, amplifies a fallback to Neutral (Fig. \ref{fig4}). Aggregating perturbation outcomes yields a stable hierarchy, indicating a strong asymmetry in how emotion representations are utilized under uncertainty. In Gemma 3n, Neutral likewise occupies the top position in the hierarchy, but the structure beneath it differs substantially from Qwen2.5, reflecting differences in pretraining and architectural design. Notably, Gemma 3n exhibits a much stronger collapse toward Neutral under Voice-only input, whereas this tendency is largely suppressed when Face information is present. As such, this emotion-label perturbation analysis empirically reveals structured inference bias patterns that may reflect distortions in transformer dynamics at the levels of semantic network structure and self- and cross-attention mechanisms.

\subsubsection*{Modality-dependent Dominance and Cross-modal Asymmetry}

\quad Despite differences in internal hierarchy and error dynamics, both models exhibit a strikingly consistent modality-level bias pattern. Specifically, Video+Audio inputs produce error structures that are far more similar to Video-only inputs than to Audio-only inputs \citep{mullick2025_emotion}. In neither model does multimodal input meaningfully reduce error bias compared to the dominant unimodal condition.

\quad For Qwen2.5, Face+Voice errors closely track those of Face-only input, whereas Voice-only errors exhibit distinct, weaker attraction patterns. For Gemma 3n, this asymmetry is even more pronounced: although Voice-only inference shows extremely strong Neutral bias under failure, this bias is almost entirely absent from Face+Voice inference, which instead mirrors Face-only behavior. In effect, the presence of video information suppresses, rather than integrates, the bias structure induced by audio.

\quad These observations suggest that, under current multimodal fusion paradigms, adding an additional modality does not necessarily act as a corrective signal. Instead, multimodality can reinforce or lock in the dominance of a single modality, producing error structures governed by the stronger representation rather than being jointly shaped by both.

\section*{Dynamical Analysis of Multimodal Interaction Using a Physical Surrogate Model}
\subsection*{Multi-oscillator Model for Transformer Dynamics}

We develop a multi-oscillator dynamical model in which vector representations evolve according to the system dynamics defined by attention layers, pointwise feedforward operations, and layer normalization \citep{lu2019_PH6,dutta2021_PH7,geshkovski2023_PH8,geshkovski2025_PH9,bruno2025_PH10,rigollet2025_PH11} . While previous studies focused on the self-attention mechanism, we describe both self-attention and cross-attention mechanisms as follows:
{
\setlength{\abovedisplayskip}{1pt}
\setlength{\belowdisplayskip}{1pt}
\begin{align}
    \label{eq:transformer_dyn}
    \begin{split}
    \tilde{x}_i(k+1) &= V_{SA}\left(x_i(k),\{x_j(k)\}^{N_E}_{j=1};k \right)
    +V_{CA}\left( x_i(k), \{z_j\}^{N_D}_{j=1} \right), \\
    x_i(k+1) &=\mathcal{N}_k\mathcal{F}_i(\mathcal{N}_k \tilde{x}_i(k+1);k),
    \end{split}
\end{align}
}where $x_i=x_i(k)\in\mathbb{R}^d$ $(i=1,2,\ldots,N_E)$ denotes the feature vector of the input token at the $k$-th layer, $V_{SA}$ represents the interaction induced by self-attention, and $V_{CA}$ cross-attention interaction with the tokens from a distinct modality $z_i\in\mathbb{R}^{d'}$ $(i=1,2,\ldots,N_D)$. Here, $\mathcal{N}_k$ denotes the layer normalization operation, typically implemented via root mean square normalization, and $\mathcal{F}_i(\,\cdot\,;k)$ the feedforward transformation at the $k$-th layer for the $i$-th token. The interaction $V_{SA}$ mediated by self-attention is defined as
{
\setlength{\abovedisplayskip}{1pt}
\setlength{\belowdisplayskip}{-2pt}
\begin{align}
    \label{eq:transformer_SA}
    V_{SA}\left(x_i(k),\{x_j(k)\}^{N_E}_{j=1}\right)
    = W_O(k)\cdot \bigoplus_{h=1}^{N_H} \left[
    \sum_{j=1}^{N_E}{\frac{e^{\langle W_Q^{(h)}(k)x_i,W^{(h)}_K(k)x_j \rangle/\sqrt{\frac{d}{N_H}}}}
    {\sum_{l=1}^{N_E}{e^{\langle  W^{(h)}_Q(k)x_i, W^{(h)}_K(k)x_l\rangle/\sqrt{\frac{d}{N_H}}}}}
    } W^{(h)}_V(k)x_j
    \right],
\end{align}
}with $W^{(h)}_Q(k)$, $W^{(h)}_K(k)$, $W^{(h)}_V(k)$ denoting the query, key and value projection matrices for the $h$-th attention head at $k$-th layer, respectively, $W_O(k)$ the output projection matrix, and $\oplus$ the concatenation (direct sum) over the $N_H$ attention heads. The cross-attention interaction $V_{CA}$ is defined by replacing the same-modality tokens with the tokens from a distinct modality ${z_i}_{i=1}^{N_D}$:
{
\setlength{\abovedisplayskip}{0pt}
\setlength{\belowdisplayskip}{-2pt}
\begin{align}
    \label{eq:transformer_CA}
    V_{CA} \left( x_i(k), \{ z_j \}_{j=1}^{N_D} \right) = 
    W'_O(k) \cdot \bigoplus_{h=1}^{N_H} \left[
    \sum_{j=1}^{N_D} \frac{e^{\langle {W'}_Q^{(h)}(k)x_i, {W'}_K^{(h)}(k)z_j \rangle / \sqrt{\frac{d}{N_H}}}}
    {\sum_{l=1}^{N_D} e^{\langle {W'}_Q^{(h)}(k)x_i, {W'}_K^{(h)}(k) z_l \rangle / \sqrt{\frac{d}{N_H}}}}
    {W'}_V^{(h)}(k)z_j
    \right],
\end{align}
}where ${W'}_Q^{(h)}$, ${W'}_K^{(h)}$, ${W'}_V^{(h)}$, ${W'}_O^{(h)}$ are query, key, value and output projection matrices for cross-attention, respectively. Within this framework, the resulting transformer dynamics constitute a complex dynamical system characterized by heterogeneous nonlinear interactions, under which emergent behavior can arise.
\begin{figure}[t]
\centering
\includegraphics[scale=0.13,angle=0]{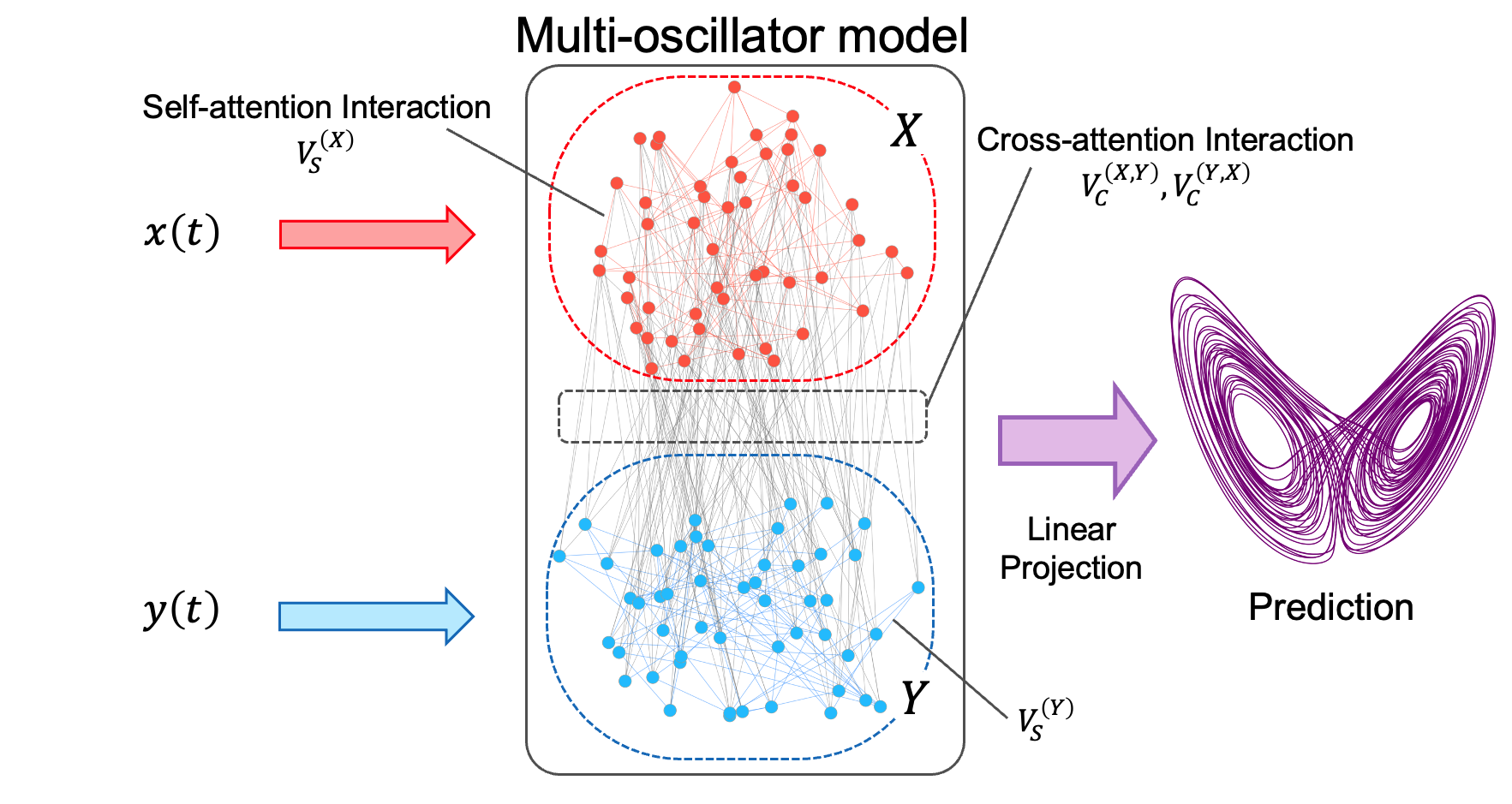}
\caption{\label{fig5}
Schematic diagram of Lorenz chaotic time-series prediction using the multi-oscillator model with self- and cross-attention mechanisms.
}
\end{figure}
The multi-oscillator model explicitly incorporates the key mechanisms of transformer architecture: self-attention and cross-attention. While previous studies have established an equivalence between the transformer dynamics and those of coupled phase oscillators under simplifying assumptions \citep{geshkovski2025_PH9} ($d=d'=2$, root mean square normalization and identity projection matrices), the present approach extends this correspondence to MLLMs by explicitly accounting for cross-modal interactions. The system consists of two groups of phase oscillators—$X$ and $Y$—each representing a distinct modality. The equation of motion of the $i$th oscillator in subgroup $\sigma \in \{X,Y\}$ is given by
{
\setlength{\abovedisplayskip}{1pt}
\setlength{\belowdisplayskip}{1pt}
\begin{equation}
    \label{eq:KM_dyn}
    \frac{d\theta_i^{(\sigma)}}{dt} = 
    V_S^{(\sigma)} \left( \theta_i^{(\sigma)}, \{ \theta_j^{(\sigma)} \}_{j=1}^{N_\sigma} \right) + 
    V_C^{(\sigma, \sigma')} \left( \theta_i^{(\sigma)}, \{ \theta_j^{(\sigma')} \}_{j=1}^{N_{\sigma'}} \right) + 
    I_i^{(\sigma)},
\end{equation}
}where $V_S^{(\sigma)}$ denotes intra-interaction within subgroup $\sigma$ and $V_C^{(\sigma,\sigma')}$ denotes inter-interaction between $\sigma$ and $\sigma'$. Here, the oscillators are driven by an external field $I^{(\sigma)}_i=h^{(\sigma)}_i\sin{\theta^{(\sigma)}_i}s_{\sigma}(t)$, with the random field strength $h^{(\sigma)}_i$ drawn from a uniform distribution over $[-h_0,h_0]$, thereby mapping information from two distinct modalities into the phase dynamics of the oscillator populations. The intra-interaction and inter-interaction terms are defined as
{
\setlength{\abovedisplayskip}{0pt}
\setlength{\belowdisplayskip}{-2pt}
\begin{align}
    \label{eq:KM_int}
    \begin{split}
    V_S^{(\sigma)} \left( \theta_i^{(\sigma)}, \{ \theta_j^{(\sigma)} \}_{j=1}^{N_\sigma} \right) &= \sum_{j=1}^{N_\sigma} K_{ij}^{(\sigma)}(t) a_{ij}^{(\sigma)} \sin \left( \theta_j^{(\sigma)} - \theta_i^{(\sigma)} \right), \\
    V_C^{(\sigma, \sigma')} \left( \theta_i^{(\sigma)}, \{ \theta_j^{(\sigma')} \}_{j=1}^{N_{\sigma'}} \right) &= \sum_{j=1}^{N_{\sigma'}} K_{ij}^{(\sigma, \sigma')}(t) a_{ij}^{(\sigma, \sigma')} \sin \left( \theta_j^{(\sigma')} - \theta_i^{(\sigma)} \right),
    \end{split}
\end{align}
}where connectivity $a_{ij}^{(\sigma)}$ and $a_{ij}^{(\sigma,\sigma')}$ encode structural constraints imposed to the semantic network. Based on observations that token representations in LLMs exhibit highly organized structures characterized by small-world topology \citep{geshkovski2023_PH8,geshkovski2025_PH9,bruno2025_PH10,liu2025new_PH12}, the connectivity is modeled using a Watts–Strogatz network with rewiring probability $p=0.01$ and degree $k=10$. Within the structural constraints, the oscillators interact with attention weights
{
\setlength{\abovedisplayskip}{0pt}
\setlength{\belowdisplayskip}{0pt}
\begin{equation}
    \label{eq:KM_attention}
    K_{ij}^{(\sigma)}(t) = K_0 \frac{e^{\beta_{\text{self}} \cos \left( \theta_i^{(\sigma)}(t) - \theta_j^{(\sigma)}(t) \right)}}{\sum_{k=1}^{N_\sigma} e^{\beta_{\text{self}} \cos \left( \theta_i^{(\sigma)}(t) - \theta_k^{(\sigma)}(t) \right)}}, 
    K_{ij}^{(\sigma, \sigma')}(t) = K_0 \frac{e^{\beta_{\text{cross}} \cos \left( \theta_i^{(\sigma)}(t) - \theta_j^{(\sigma')}(t) \right)}}{\sum_{k=1}^{N_{\sigma'}} e^{\beta_{\text{cross}} \cos \left( \theta_i^{(\sigma)}(t) - \theta_k^{(\sigma')}(t) \right)}},
\end{equation}
}where $\beta_\text{self}$ and $\beta_\text{cross}$ determine the strength of self- and cross-attention, respectively. In the limit $\beta_\text{self}, \beta_\text{cross} \rightarrow 0$, the attention weights reduce to uniform coupling—the attention mechanism is effectively switched off. 

\quad To investigate the dynamics of biased modality preferences, we perform the Lorenz chaotic time-series prediction task. The Lorenz system was initially introduced to model atmospheric convection \citep{lorenz1963_PH13}, which was the first demonstration of chaos in dynamical systems, widely known as the butterfly effect, whereby small perturbations can cause significant, unpredictable changes in system evolution. It has served as a canonical model of chaotic dynamics across various fields, including quantum optics \citep{elgin1987_PH14}, fluid dynamics \citep{takens1981_PH15}, cryptography \citep{kocarev1995_PH16,lin2021_PH17}, finance \citep{olkhov2018_PH18}, and social sciences \citep{kiel1997_PH19}. 

\quad As shown in Figure \ref{fig5}, oscillators in subgroup $X$ are driven by the $x$-component of the Lorenz system, $s_X (t)=x(t)$, while oscillators in $Y$ are driven by the $y$-component, $s_Y (t)=y(t)$.  The z-component z(t) can be inferred from the oscillator observables $\{ \sin{\theta^{(\sigma)}}\}^{\sigma\in\{X,Y\}}_{i=1,2,\ldots,N_\sigma}$ via a linear projection. The projection weights are obtained using ridge regression between the observables and the target $z(t)$ over the interval $0\leq t \leq 30$, and the prediction accuracy is evaluated by the normalized mean squared error (NMSE) over $30<t\leq80$. A dynamical SHAP is defined to quantify the contribution of each input component ($x(t)$ or $y(t)$) in the prediction of $z(t)$. The dynamical SHAP value of each modality is defined as $\phi(\sigma)=[-(\text{NMSE}_{X\cup Y} - \text{NMSE}_\sigma) - (\text{NMSE}_\sigma - \text{NMSE}_{\emptyset})]/2$, where $\text{NMSE}_S$ denotes NMSE for the prediction using oscillator observables in group $S$. The difference in dynamical SHAP value between the two modalities, $\phi(Y)-\phi(X)$. The base line error for the null set is given by $\text{NMSE}_{\emptyset}=1$. A larger $\phi(\sigma)$ indicates a stronger contribution of modality $\sigma$ to prediction performance. A positive value of $\phi(Y)-\phi(X)$ thus signifies dominance of modality $Y$, while a negative value indicates dominance of $X$.

\begin{figure}[t!]
\centering
\includegraphics[width = \textwidth]{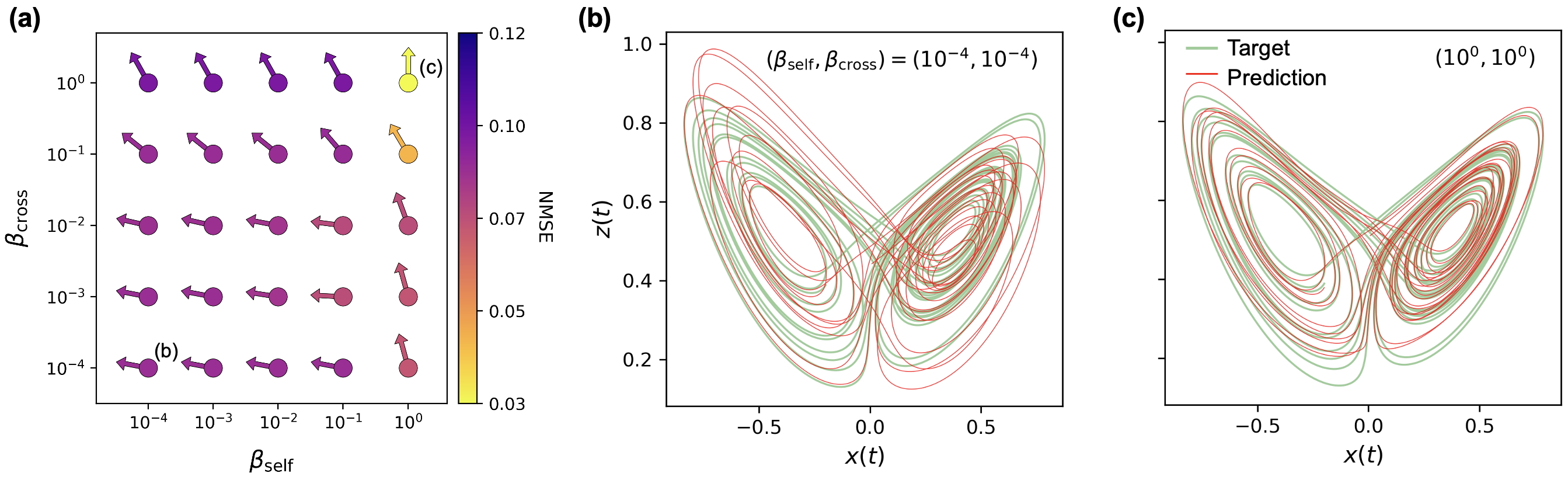}
\caption{\label{fig6}
Characterization of the transformer dynamics using a physical testbed: Lorenz chaotic time-series prediction on a multi-oscillator system. (a) Modality preference is quantified by the difference in dynamical SHAP values, $\phi(Y)-\phi(X)$, across the self- and cross-attention levels $(\beta_\text{self},\beta_\text{cross})$. The SHAP difference is represented by the direction of the arrow in the range of $[-90^\circ,90^\circ]$: $0^\circ$ signify the equal contribution of $X$ and $Y$, $-90^\circ$ indicate the $X$-only; $90^\circ$ the $Y$-only. The arrow color represents the normalized mean squared error (NMSE) between target $z(t)$ and prediction. Predictions are visualized in the embedding space for two representative cases: (b) low self- and cross-attention levels, i.e., $(\beta_\text{self},\beta_\text{cross})= (10^{-4},10^{-4})$ and (c) high levels $(\beta_\text{self},\beta_\text{cross})= (10^{0},10^{0})$. To clarify, only the time series for $50\leq t \leq70$ are displayed.}
\end{figure}

\quad The system consists of $N_X=50$ and $N_Y=50$ oscillators, which evolve according to the multimodal dynamics in the equations (4-6) with $K_0=2(N_X+N_Y)$ and $h_0=0.5$. Figure \ref{fig6}(a) displays the modality preference $\phi(Y)-\phi(X)$ with the prediction error NMSE ($=\text{NMSE}_{X\cup Y}$), across the self- and cross-attention levels. For low self ($\beta_\text{self}$)- and cross-attention ($\beta_\text{cross}$) levels, the prediction is largely dominated by X, as indicated by the left-directing arrows for the SHAP difference (Fig. \ref{fig6}a). In this case the prediction error is relatively high (Fig. \ref{fig6}b). As both $\beta_\text{self}$ and $\beta_\text{cross}$ increase, the contributions of X and Y to inference become similar. With sufficiently high attention levels, i.e., $(\beta_\text{self},\beta_\text{cross})= (10^{0},10^{0})$, the contributions from the two modalities are equal, $\phi(X)\approx \phi(Y)$, where the inference accuracy manifests the highest value. In this case, the prediction well reproduces the attractor structure in the embedding space (Fig. \ref{fig6}c). 

\section*{Conclusions and Remarks}

The present paper explores both empirical evidence for popular MLLMs (Qwen2.5-Omni and Gemma 3n) and the dynamical principles underlying the distortions in transformer dynamics that give rise to biased modality preferences in multimodal inference. Although a complete understanding of how this multimodal bias arises and how it can be controlled remains for future research, the results obtained indicate that adequate self- and cross-attention levels are crucial for preventing multimodal bias in transformer dynamics (presumably at the levels of embedding, latent, and representation), promoting balanced use of multimodal inputs, and ultimately enhancing computational capabilities. However, we do not claim that such bias in MLLMs is an inevitable consequence of multimodal learning. Rather, these results indicate a structural tendency under widely used multimodal pretraining and fusion schemes, where disparities in representation scale, token counts, temporal richness, and cross-attention normalization can plausibly bias interaction dynamics toward one modality. From a bias management perspective, this analysis highlights a failure mode that standard fairness metrics and performance evaluations are ill-equipped to detect. While multimodal models may appear competitive or even improved under aggregate metrics, their error dynamics reveal systematic dominance patterns that undermine the intended benefits of multimodal integration. The graph-based characterization employed here provides a compact and interpretable diagnostic tool. It serves as a natural bridge to the physics-based surrogate dynamical model introduced in the following section. By first establishing empirical regularities across two distinct MLLM families, we create a foundation for interpreting multimodal bias not merely as a representational artifact, but as an emergent property of cross-modal interaction dynamics.

\quad On the other hand, our physics-based method for understanding the multimodal bias is grounded in phenomenology, which emphasizes a rigorous description of lived experience but puts on hold the question of the existence of an external reality \citep{husserl2012i_CR4,merleau1969_CR5}. This approach was developed in the first half of the 20th century by Edmund Husserl, Martin Heidegger, Maurice Merleau-Ponty, and others, and is prized as the proper foundation of modern philosophy. As such, our phenomenological approach focuses on the physical internal entities that machine experiences during training or inference at embedding and representation levels, unlike cognitivist interpretations, which hold that networks encode representations of external entities. Indeed, there is ongoing debate about whether representationalism-cognitivism accounts, which trace symbolic vector representations of external reality within neural networks, can properly describe modern AI functions \citep{beckmann2023_CR6,jantzen2025_CR7}, which are heavily influenced by biology- and psychology-based traditions for understanding human intelligence, either through the localist \citep{bowers2009_CR8,minsky1969_CR9,turing1950_CR10,turing1936_CR11,newell2007_CR12} or the connectionist view \citep{anderson1931_CR13,thorndike1931_CR14,minsky1969_CR9,rosenblatt1958_CR15,mcculloch1943_CR16,lecun2015_CR17} . A physics-based, phenomenological approach rather than a cognitivist view may lay the foundations for dynamics-oriented, explainable methods to reveal the subtle distortions in transformer operations that cause systematic bias in MLLMs.

\section*{Acknowledgments}

This work was supported by Institute of Information \& communications Technology Planning \& Evaluation (IITP) grant funded by the Korea government (MSIT) (No.RS-2025-02217259, Development of self-evolving AI bias detection-correction-explanation platform based on international multidisciplinary governance, 80\%) and the Korea Institute of Science and Technology (KIST) Institutional Program (Project no. 2E33721, 10\%). J.W. was supported by Basic Science Research Program through the National Research Foundation of Korea (NRF) funded by the Ministry of Education (RS-2024-00452400, 10\%).



\bibliography{references}



\end{document}